\documentclass[runningheads]{llncs}
\usepackage[T1]{fontenc}
\usepackage{graphicx}
\usepackage{amsmath}
\usepackage{amsfonts} 
\usepackage{booktabs}
\usepackage{float}
\makeatletter
\renewcommand\subsubsection{\@startsection{subsubsection}{3}{\z@}%
                       {-18\p@ \@plus -4\p@ \@minus -4\p@}%
                       {4\p@ \@plus 2\p@ \@minus 2\p@}
                       {\normalfont\normalsize\bfseries\boldmath}}
\makeatother
\begin{document}
%
\title{OPTNet: Ordering Point Transformer Network for Post-disaster 3D Semantic Segmentation}
\titlerunning{OPTNet}
%
\author{Nhut Le \inst{1} \and
Ehsan Karimi\inst{1} \and
Maryam Rahnemoonfar\inst{1,2, \thanks{Corresponding Author: Maryam Rahnemoonfar. Email: maryam@lehigh.edu}}}
\authorrunning{N. Le et al.}
%
\institute{Computer Science and Engineering, Lehigh University, Bethlehem PA 18015, US \and
Civil and Environmental Engineering, Lehigh University, Bethlehem PA 18015, US \\
\email{\{nhl224, ehk224, maryam\}@lehigh.edu}}
\maketitle              
%
\begin{abstract}
Post-disaster damage assessment requires rapid and accurate semantic segmentation of 3D point clouds to identify critical infrastructure such as damaged buildings and roads. Early Point Transformers (e.g., PTv1, PTv2) relied on computationally expensive neighbor searching (k-NN) and Farthest Point Sampling (FPS). To improve efficiency, recent architectures like Point Transformer V3 (PTv3) adopted static serialization methods, such as Hilbert curves or Z-order, to organize unstructured points for window-based attention. However, these fixed orderings are not optimal for capturing the complex geometry of disaster scenes. In this paper, we propose OPTNet (Ordering Point Transformer Network), which introduces a learnable Point Sorter module. OPTNet utilizes a self-supervised ordering loss to dynamically predict an optimal permutation that maximizes the locality of the attention mechanism. We evaluate our method on the 3DAeroRelief dataset, significantly outperforming state-of-the-art baselines.

\end{abstract}    
\section{Introduction}
\label{sec:intro}
\begin{figure}[t]
    \centering
    \includegraphics[width=\linewidth]{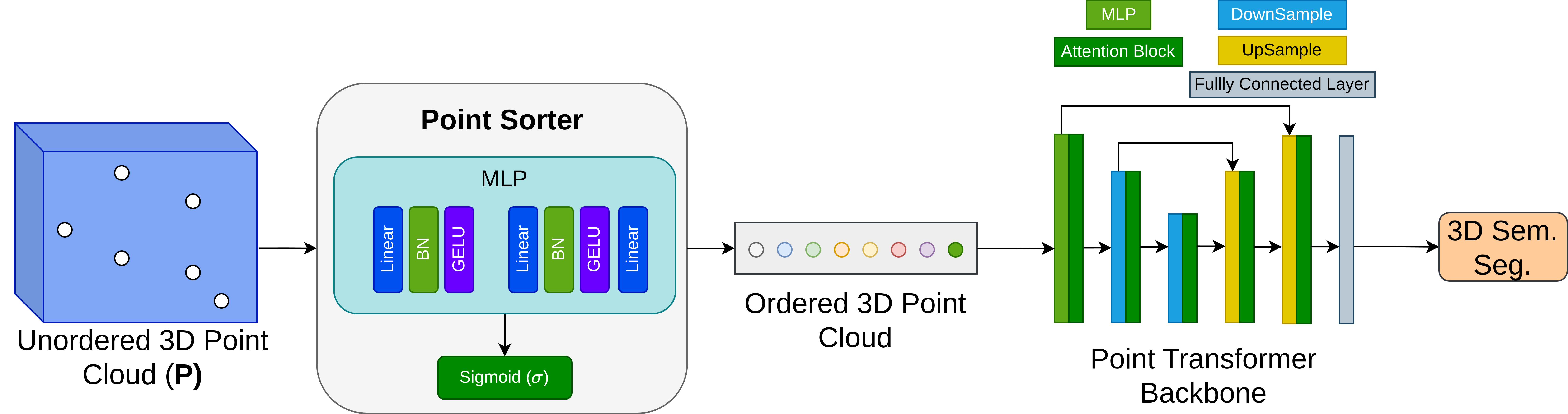}
    \caption{\textbf{The overview of OPTNet Framework.} The network utilizes a Learnable Point Sorter to dynamically serialize the input point cloud, optimizing the point order for the subsequent Point Transformer Backbone. This learnable serialization preserves geometric locality more effectively than static heuristics, enhancing the efficiency of windowed attention.}    
\label{fig:framework}
\end{figure}

3D semantic segmentation is a cornerstone of scene understanding for robotics, autonomous driving, and augmented reality. To process irregular 3D point clouds effectively, recent state-of-the-art approaches have adopted diverse structuring strategies. Classical and hierarchical methods \cite{qi2016pointnet,qi2017pointnetplusplus,qian2022pointnext,zhao2021point,wu2022point} rely on explicit neighbor grouping (e.g., $k$-NN or ball query) and Farthest Point Sampling (FPS), which can be computationally expensive and difficult to scale. Voxel-based sparse convolutional networks \cite{choy20194d,park2022fast,robert2023spt} address this with efficient hashing but often lack the global receptive field of transformers. Consequently, the field has recently shifted toward serialized point transformers \cite{wu2024ptv3}, which map 3D points into 1D ordered sequences using heuristic space-filling curves (e.g., Morton or Hilbert codes). This serialization enables highly efficient, contiguous windowed attention without the latency of complex grouping or hash map construction.
\\
However, the efficacy of this serialization strategy depends entirely on the quality of the sorting order. Static, heuristic space-filling curves impose a fixed traversal pattern with inherent boundary artifacts - points that are spatially adjacent in 3D can end up far apart in the 1D sequence, shattering the local receptive field required for effective windowed attention.
\\
We argue that the optimal serialization order for 3D semantic segmentation should not be fixed a priori, but rather learned alongside the features. To this end, we present the Ordering Point Transformer Network (OPTNet), a novel framework that dynamically optimizes point serialization to maximize the efficiency of windowed transformers (see figure \ref{fig:framework}). By replacing static heuristics with a content-aware mechanism, OPTNet preserves the speed benefits of serialization while overcoming the geometric limitations of fixed curves. 
\\
The core innovation of OPTNet is the Point Sorter, a lightweight module that predicts a sorting score for each point based on its geometry and features. To train this scorer effectively without manual annotation of "optimal" order, we introduce a Self-Supervised Ordering Loss. Unlike previous methods that might try to mimic specific curves, we directly optimize for the properties of a good sequence: Locality and Uniformity. The Locality Loss enforces that points which are neighbors in 3D space receive similar scores, ensuring they remain adjacent in the serialized sequence. Simultaneously, the Distribution Loss enforces that the predicted scores span the entire $[0,1]$ range uniformly, preventing mode collapse where all points are mapped to a single value.
\\
Leveraging this learned order, OPTNet feeds the optimized sequence into a standard serialized transformer backbone. By sorting points according to the predicted scores, we ensure that semantically and spatially related points are grouped together. This allows the subsequent attention layers to capture complex local neighborhoods effectively, adapting to the underlying geometry without requiring changes to the efficient serialization-based architecture.

\subsection{Overview of OPTNet} The pipeline comprises three key stages (see figure \ref{fig:framework}):
\begin{enumerate}
    \item \textbf{Learnable Point Sorter} the Point Sorter is a lightweight multi-layer perceptron (MLP). It predicts a scalar value for each point. Sorting the point cloud by these scores generates a 1D sequence optimized for locality.
    \item \textbf{Self-Supervised Optimization.} During training, the scorer is supervised via a dual-term loss function. The Locality Loss minimizes score variance among spatial neighbors, while the Distribution Loss prevents collapse by matching the score distribution to a uniform prior.
    \item \textbf{Serialized Transformer Processing.} The network processes the ordered sequence using efficient windowed attention blocks. The learnable order ensures that simple contiguous windows capture relevant geometric features, avoiding the boundary artifacts common in static sorting methods.
\end{enumerate}

\subsection{Contributions} This work makes the following contributions:
\begin{enumerate}
    \item \textbf{OPTNet Framework:} A streamlined approach that enhances point transformers by replacing static sorting with a learnable serialization mechanism, eliminating the need for complex grouping operations (FPS/KNN).
    \item \textbf{Self-Supervised Ordering:} A novel training paradigm that optimizes geometric ordering through Locality and Distribution constraints, enabling the network to learn a robust, artifact-free traversal order without heuristic supervision.
    \item \textbf{Seamless Integration:} We demonstrate that our learnable sorter can directly replace static space-filling curves in existing transformer architectures, enhancing their ability to capture complex topologies.
    \item \textbf{Extensive Experiment:} We demonstrate that OPTNet achieves superior accuracy-efficiency trade-offs, exceeding state-of-the-art results on a post-disaster 3D dataset - 3DAeroRelief \cite{le20253daerorelief3dbenchmarkuav}.
\end{enumerate}
\section{Related Work}
\label{sec:related}
In this section, we review the existing literature relevant to our work. We first discuss 2D semantic segmentation methods for disaster assessment, followed by an overview of 3D semantic segmentation approaches, including point-based, voxel-based, and serialization-based methods. Finally, we highlight the emerging transition to 3D-based damage assessment.
\subsection{2D Semantic Segmentation for Natural Disaster}
Rapid damage assessment using 2D imagery, primarily from satellites and Unmanned Aerial Vehicles (UAVs), has been extensively studied. Early approaches utilized Convolutional Neural Networks (CNNs) such as U-Net, PSPNet, and DeepLabv3+ to segment critical infrastructure like buildings and roads from post-disaster aerial footage \cite{rahnemoonfar2021floodnet,rahnemoonfar2023rescuenet}. 
To address the complexity of damage classification, many methods adopt a two-stage pipeline: first localizing objects (e.g., buildings) and then classifying their damage severity \cite{gupta2019creating}. 
For more granular assessment, change detection techniques employing Siamese networks \cite{chen2022learning} compare pre- and post-disaster image pairs to identify structural changes. 
Recently, vision foundation models like the Segment Anything Model (SAM) have been adapted for disaster scenarios, using visual prompting or learnable adapters to handle domain shifts and limited labeled data \cite{li2024visual}. 
However, 2D-based methods inherently lack depth information that makes it difficult to distinguish between certain damage types (e.g., pancaked buildings vs. intact roofs) or handle severe occlusions. Thus, it necessitates the move to 3D representations.
\subsection{3D Semantic Segmentation for Natural Disaster}
Recently, the field has begun transitioning from 2D to 3D assessment to leverage geometric information. Le et al. \cite{le2025spie} introduced the first 3D semantic segmentation network specifically designed for post-disaster assessment. However, their method was evaluated on a small-scale 3D benchmark, and the segmentation performance remains suboptimal for complex disaster scenes. Our proposed OPTNet significantly outperforms this baseline by learning dynamic ordering strategies that better capture the irregular geometry of damaged infrastructure.
\\
Deep learning approaches for general 3D semantic segmentation fall into four categories: point-based, voxel-based, graph-based, and serialization-based methods.
\subsubsection{Point, Voxel, and Graph Methods} 
Point-based methods process raw point clouds to preserve geometric details. Early MLP-based architectures \cite{qi2016pointnet,qi2017pointnetplusplus,qian2022pointnext} introduced hierarchical feature learning but relied on computationally expensive FPS and $k$-NN grouping. Convolutional variants \cite{NEURIPS2018_f5f8590c,Thomas_2019_ICCV} adapted standard operations to irregular data, while RandLA-Net~\cite{hu2019randla} utilized random sampling for efficiency, often at the cost of geometric fidelity. Recently, OA-CNNs~\cite{Peng_2024_CVPR} improved performance via adaptive receptive fields but still face scalability challenges.
Alternatively, voxel-based methods \cite{choy20194d,robert2023spt,yang2023swin3d,park2022fast} convert points into regular grids to leverage sparse convolutions, introducing quantization artifacts. Graph-based approaches \cite{Landrieu2017-hv,robert2023spt} construct complex graphs of geometrically homogeneous regions to model long-range dependencies.
\subsubsection{Point Transformers and Serialization} 
Attention-based models have become dominant for their global context capabilities. Early iterations \cite{zhao2021point,wu2022point} utilized vector attention but remained bottlenecked by explicit neighbor grouping. To address this, Point Transformer v3 (PTv3)~\cite{wu2024ptv3} pioneered serialization-based processing, mapping irregular 3D points into a 1D structured sequence using static space-filling curves (e.g., Hilbert or Z-order). This serialization enables highly efficient contiguous windowed attention. However, PTv3~\cite{wu2024ptv3} relies on fixed, heuristic ordering strategies that impose a rigid traversal pattern, creating boundary artifacts where spatially adjacent points are separated in the sequence. \\
OPTNet builds upon the efficiency of the serialization paradigm established by PTv3~\cite{wu2024ptv3} but departs from static heuristics. Instead of relying on fixed space-filling curves, we introduce a learnable Point Sorter trained via a self-supervised ordering loss. This allows the network to dynamically optimize the traversal order based on the specific geometry of the scene. By optimizing for locality and uniform distribution, OPTNet enhances the feature aggregation of the transformer backbone without requiring complex modifications to the underlying architecture.

\section{OPTNet: Ordering Point Transformer Network}
\label{sec:method}

Let a 3D scene be represented as a point set $\mathcal{P}=\{p_i\}_{i=1}^{N}$ with associated features $\mathbf{f}_i \in \mathbb{R}^{F}$, comprising coordinates and color/intensity. The goal of 3D semantic segmentation is to assign a semantic label $y_i \in \{1,\dots,C\}$ to every point.
\\
To process large-scale point clouds efficiently, OPTNet adopts a serialization-based architecture.
As illustrated in figure \ref{fig:framework} and figure \ref{fig:pointsorter}, our framework consists of three main components: a Learnable Point Sorter that dynamically orders the points, a Point Transformer Backbone for hierarchical feature learning, and a Self-Supervised Ordering Loss to guide the sorting process in training.
\subsection{Point Transformer Backbone}
The Point Transformer Backbone is responsible for extracting semantic features from the serialized point cloud. It follows a U-Net style Encoder-Decoder architecture composed of four key operations:
\begin{itemize}
    \item \textbf{Input Embedding.}
The raw 3D point cloud (coordinates and features) is first passed through a MLP to project the inputs into a high-dimensional feature space, creating the initial embeddings.
    \item \textbf{Serialized Attention Block.}
The core computation unit is the windowed self-attention block.
Using the 1D order provided by the Point Sorter, points are grouped into local windows.
Multi-head self-attention is applied within each window to aggregate context.
This serialization transforms the complexity of finding neighbors from $O(N \log N)$ in 3D space to $O(N)$ in the 1D sequence.
\item \textbf{Downsampling (Encoder).}
The encoder consists of multiple stages.
Between stages, we employ a Downsampling layer (Grid Pooling).
This operation merges points within the same grid cell to reduce the spatial resolution while increasing the feature channel dimension.
\item \textbf{Upsampling (Decoder).} The decoder also consists of multiple stages, mirroring the encoder. We use an Upsampling layer (Linear Interpolation) to restore the spatial resolution.
Features from the encoder are fused via skip connections to preserve fine-grained geometric details.
\end{itemize}
\noindent

\subsection{Point Sorter Module}
\label{sec:sorter}
\begin{figure}[!t]
    \centering
    \includegraphics[width=\linewidth]{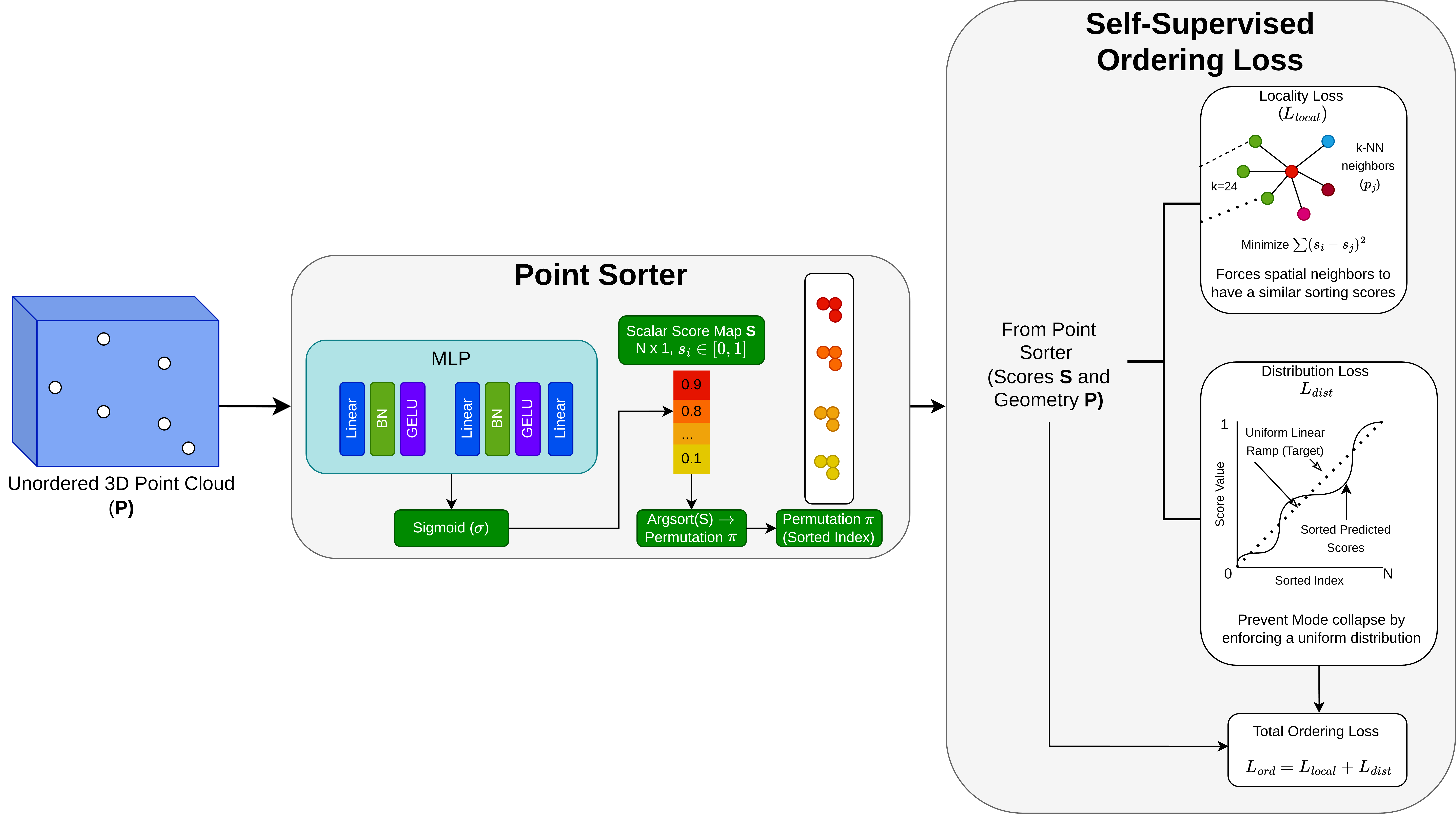}
    \caption{\textbf{The core mechanism of OPTNet.}
\textit{Point Sorter:} An MLP consumes point coordinates and features to predict a scalar score $s_i \in [0,1]$ for every point. These scores are sorted to produce a permutation $\pi$ that serializes the point cloud.
\textit{Self-Supervised Ordering Loss:} We train the sorter using a Locality Loss ($\mathcal{L}_{local}$), which minimizes the score variance among spatial $k$-nearest neighbors to preserve geometric structure, and a Distribution Loss ($\mathcal{L}_{dist}$), which forces the scores to follow a uniform distribution to prevent mode collapse.}
    \label{fig:pointsorter}
\end{figure}
The core component of OPTNet is the Point Sorter, a lightweight MLP designed to predict a scalar sorting score for each point (see figure \ref{fig:pointsorter}).
Given a point $p_i$ with coordinates $\mathbf{x}_i$ and features $\mathbf{h}_i$, we concatenate them to form the input vector. The scorer $f_\theta$ maps this input to a scalar score $s_i \in [0, 1]$:

\begin{equation}
    s_i = \sigma(\text{MLP}(\text{concat}(\mathbf{x}_i, \mathbf{h}_i)))
\end{equation}
where $\sigma$ is the sigmoid function. The MLP consists of three linear layers with Batch Normalization and GELU activation.
We then compute the permutation $\pi$ that sorts the points based on these learned scores: $\pi = \mathrm{argsort}(\{s_i\}_{i=1}^{N})$. The reordered sequence is then fed into the subsequent transformer blocks.

\subsection{Self-Supervised Ordering Loss}
\label{sec:loss}
Training a scorer to discover a good ordering is non-trivial because there is no single "ground truth" optimal permutation. Instead of distilling from static teachers (which would just replicate their flaws), we propose a Self-Supervised Ordering Loss that directly optimizes for the desired properties of a good serialization: Locality and Uniformity (see figure \ref{fig:pointsorter}).

\subsubsection{Locality Loss ($\mathcal{L}_{local}$)}
The fundamental requirement for windowed attention is that points that are close in 3D space should be close in the 1D sequence. We enforce this by minimizing the variance of scores within local 3D neighborhoods. For each point $i$, we identify its $k$ nearest neighbors $\mathcal{N}_k(i)$ in Euclidean space and penalize the difference between their scores:

\begin{equation}
    \mathcal{L}_{local} = \frac{1}{N} \sum_{i=1}^{N} \sum_{j \in \mathcal{N}_k(i)} (s_i - s_j)^2
\end{equation}

This loss forces the scorer to assign similar rank values to geometrically adjacent points, ensuring they remain clustered in the sorted sequence.

\subsubsection{Distribution Loss ($\mathcal{L}_{dist}$)}
Minimizing $\mathcal{L}_{local}$ alone can lead to a trivial solution where $s_i = \text{const}$ for all $i$ (mode collapse). To prevent this, we impose a constraint that the predicted scores must be uniformly distributed across the interval $[0, 1]$. We achieve this by sorting the predicted scores vector $\mathbf{s}$ and penalizing its deviation from a fixed uniform linear ramp $\mathbf{t}$, where $t_i = i/N$:

\begin{equation}
    \mathcal{L}_{dist} = \frac{1}{N} \sum_{i=1}^{N} (\text{sort}(\mathbf{s})_i - t_i)^2
\end{equation}

This forces the network to utilize the full dynamic range of sorting scores, ensuring a valid and balanced permutation. The total ordering loss is $\mathcal{L}_{ord} = \mathcal{L}_{local} + \mathcal{L}_{dist}$.

\subsection{Training Strategy}
\begin{itemize}
    \item \textbf{Warmup Phase:} To stabilize training, we employ a warmup strategy for the first $W$ epochs. During this phase, we bypass the learned permutation and use a static space filling curves (the shuffle of Z-order, Hilbert and their inverse) to serialize the points. This allows the transformer weights to converge to a reasonable state using a known good ordering. Simultaneously, the Point Sorter is trained via $\mathcal{L}_{ord}$ to predict meaningful scores.
    \item \textbf{Joint Training:} After epoch $W$, we switch to the learned permutation $\pi$. The entire network is trained end-to-end using a combined loss:
\begin{equation}
    \mathcal{L}_{total} = \mathcal{L}_{seg} + \lambda \mathcal{L}_{ord}
\end{equation}
where $\mathcal{L}_{seg}$ is the standard segmentation loss (Cross Entropy + Lovasz) and $\lambda$ is the weight for the ordering auxiliary task.
\end{itemize}

\section{Experiment}
\label{sec:exp}
\subsection{Dataset}
We validate our proposed method on the 3DAeroRelief dataset \cite{le20253daerorelief3dbenchmarkuav}, a specialized benchmark tailored for 3D post-disaster scene understanding.
Collected in the wake of Hurricane Ian (2022) in Florida, this dataset consists of high-fidelity 3D point clouds reconstructed from aerial UAV imagery using Structure-from-Motion (SfM) and Multi-View Stereo (MVS) pipelines.
The dataset provides dense point-wise semantic labels across five categories: \textit{Background, Tree, Road, Building-No-Damage,} and \textit{Building-Damage}.
A unique characteristic of 3DAeroRelief is its focus on structural integrity, explicitly separating damaged infrastructure from intact buildings, which is critical for emergency response applications.
For evaluation, we adhere to the official cross-area data split to ensure rigorous testing of generalization.
The training set encompasses 56 point clouds spanning seven distinct geographic zones (Areas 1, 3, 4, 5, 6, 7, and 8).
The test set is composed of 8 point clouds from a reserved, unseen location (Area 2), challenging the model to adapt to new environments not encountered during training.
\subsection{Evaluation Metrics}
To comprehensively evaluate the segmentation performance, we report three standard metrics: Mean Intersection over Union (mIoU), Mean Accuracy (mAcc), and Overall Accuracy (OA).
\begin{itemize}
    \item \textbf{Mean Intersection over Union (mIoU):} The primary metric for semantic segmentation, calculated as the average of the Intersection over Union (IoU) for each class.
    IoU measures the overlap between the predicted and ground truth points:
    \begin{equation}
        \text{mIoU} = \frac{1}{C} \sum_{c=1}^{C} \frac{\text{TP}_c}{\text{TP}_c + \text{FP}_c + \text{FN}_c}
    \end{equation}
    where $C$ is the number of classes, and $\text{TP}_c$, $\text{FP}_c$, and $\text{FN}_c$ represent the true positives, false positives, and false negatives for class $c$, respectively.
    \item \textbf{Mean Accuracy (mAcc):} The average of the class-wise accuracies (Recall), representing the model's ability to correctly classify points within each specific category, independent of class imbalance.
    \item \textbf{Overall Accuracy (OA):} The ratio of correctly classified points to the total number of points in the test set.
\end{itemize}
\begin{table}[!t]
\centering
\caption{Quantitative comparison of IoU (\%) on the 3DAeroRelief dataset \cite{le20253daerorelief3dbenchmarkuav} The best results are highlighted in \textbf{bold}.}
\label{tab:iou_results}
\begin{tabular}{l|c|c|c|c|c|c}
\toprule
\textbf{Method} & \textbf{mIoU} & \textbf{Background} & \textbf{Bldg-Dmg} & \textbf{Bldg-No-Dmg} & \textbf{Road} & \textbf{Tree} \\
\midrule
PTv1 \cite{wu2022point}     & 30.84 & 38.54 & 51.12 & 15.43 & 30.02 & 19.09 \\
PTv2 \cite{wu2022point}     & 25.59 & 50.14 & 0.20  & 25.31 & 48.64 & 3.67 \\
OA-CNNs \cite{Peng_2024_CVPR}& 26.04 & 22.77 & 58.22 & 19.62 & 0.00 & 29.59 \\
PTv3 \cite{wu2024ptv3}      & 45.84 & 77.30 & 76.14 & 31.08 & 4.67 & 40.01 \\
\midrule
\textbf{OPTNet (Ours)} 
& \textbf{79.65} & \textbf{92.40} & \textbf{94.60} & \textbf{77.08} & \textbf{51.31} & \textbf{82.85} \\
\bottomrule
\end{tabular}
\end{table}
\begin{table}[!ht]
\centering
\caption{Quantitative comparison of Class Accuracy (Acc \%)  on the 3DAeroRelief dataset \cite{le20253daerorelief3dbenchmarkuav}.
OA denotes Overall Accuracy.}
\label{tab:acc_results}
\resizebox{\textwidth}{!}{
\begin{tabular}{l|c|c|c|c|c|c|c}
\toprule
\textbf{Method} & \textbf{mAcc} & \textbf{OA} & \textbf{Background} & \textbf{Bldg-Dmg} & \textbf{Bldg-No-Dmg} & \textbf{Road} & \textbf{Tree} \\
\midrule
PTv1 \cite{wu2022point}      & 46.06 & 58.27 & 42.97 & 76.50 & 30.10 & 30.02 & 50.71 \\
PTv2 \cite{wu2022point}      & 39.41 & 31.01 & 60.77 & 0.20  & 48.26 & 51.52 & 36.28 \\
OA-CNNs \cite{Peng_2024_CVPR}& 43.96 & 58.78 & 24.27 & 92.76 & 30.30 & 0.00  & 72.48 \\
PTv3 \cite{wu2024ptv3}       & 55.08 & 83.66 & 87.34 & 89.49 & 47.97 & 47.97 & 41.49 \\
\midrule
\textbf{OPTNet 
(Ours)}       & \textbf{92.14} & \textbf{95.75} & \textbf{94.82} & \textbf{97.47} & \textbf{89.63} & \textbf{84.80} & \textbf{93.99} \\
\bottomrule
\end{tabular}
}
\end{table}
{
\setlength{\tabcolsep}{0.5pt}
\renewcommand{\arraystretch}{0.05}
\begin{table}[H]
\caption{\textbf{Qualitative Results.} Visual comparison of OA-CNNs\cite{Peng_2024_CVPR}, PTv3 \cite{wu2024ptv3}, and OPTNet (Ours) on the 3DAeroRelief test set. Building-No-Damage (Cyan), Building-Damage (Red), Tree (Green), Road (Yellow), and Background (Black).}
\centering
\resizebox{\textwidth}{!}{
\begin{tabular}{cccc}
 & \includegraphics[width=0.2\linewidth]{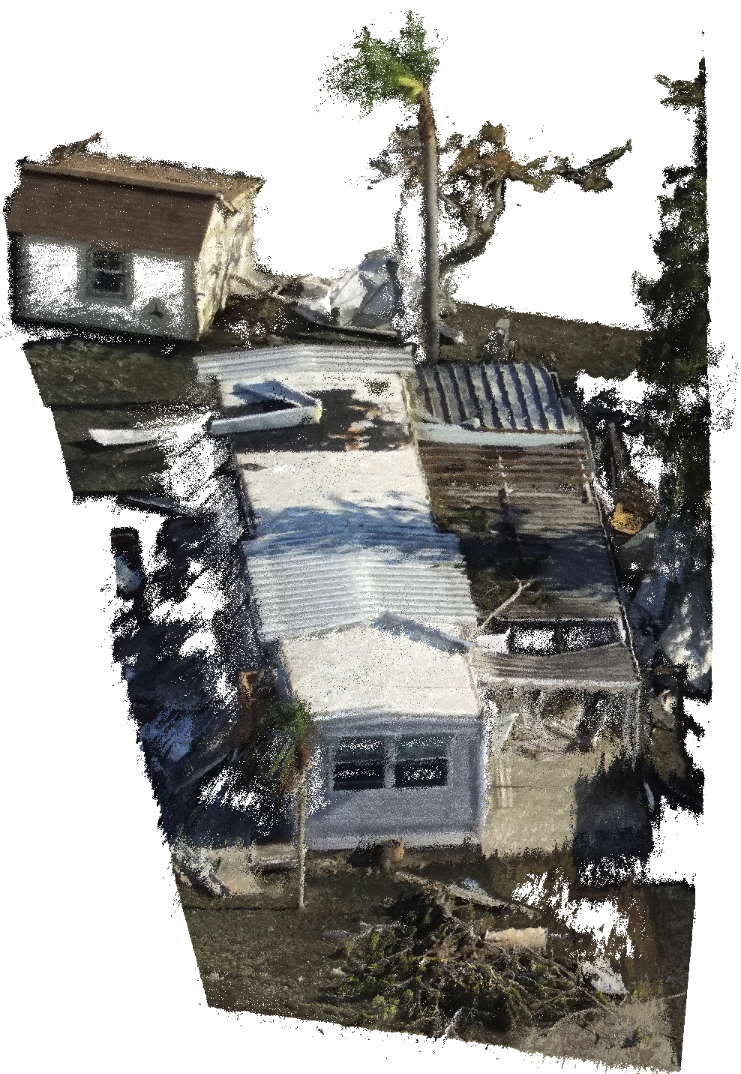}  & \includegraphics[width=0.3\linewidth]{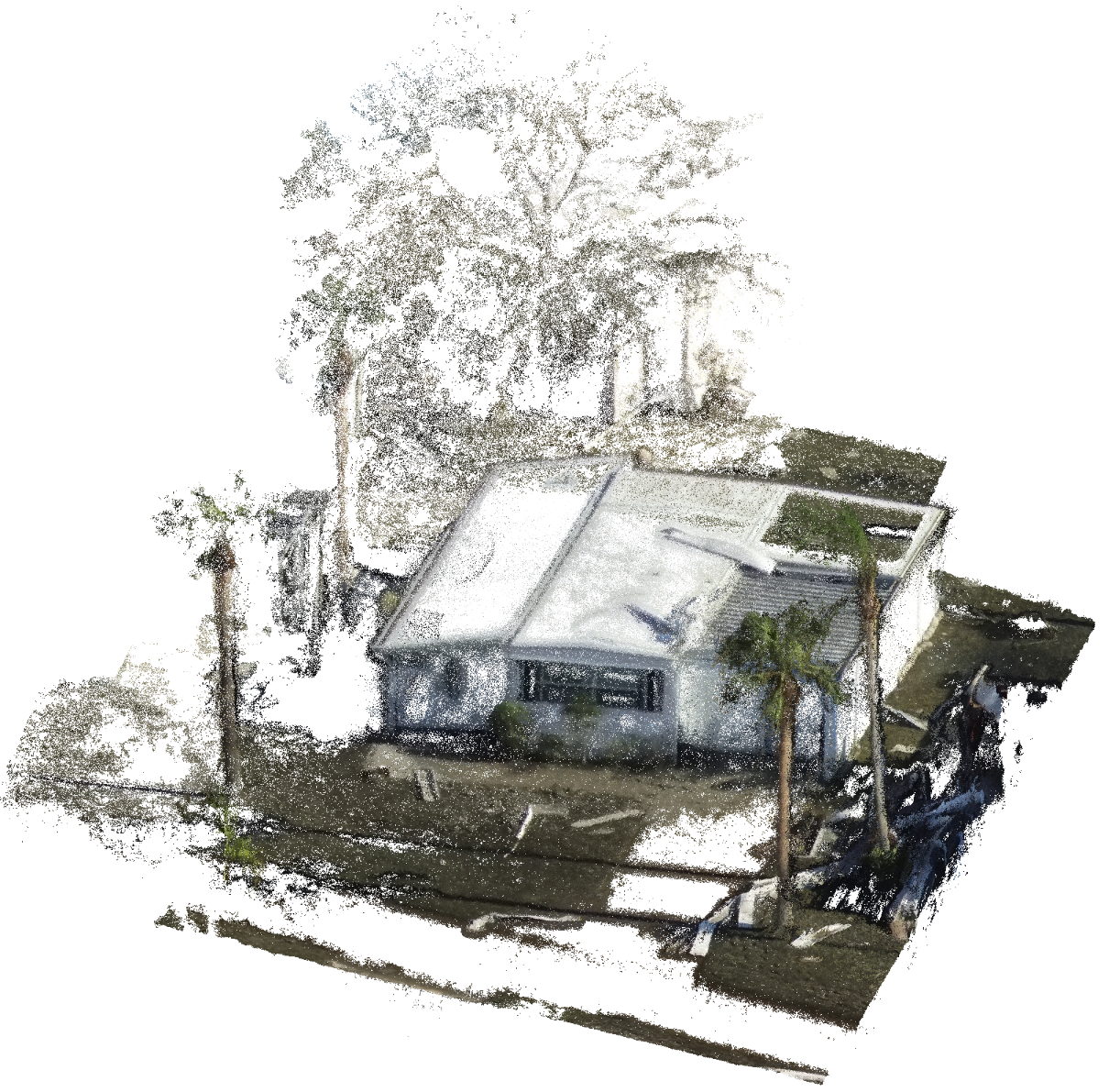} & \includegraphics[width=0.23\linewidth]{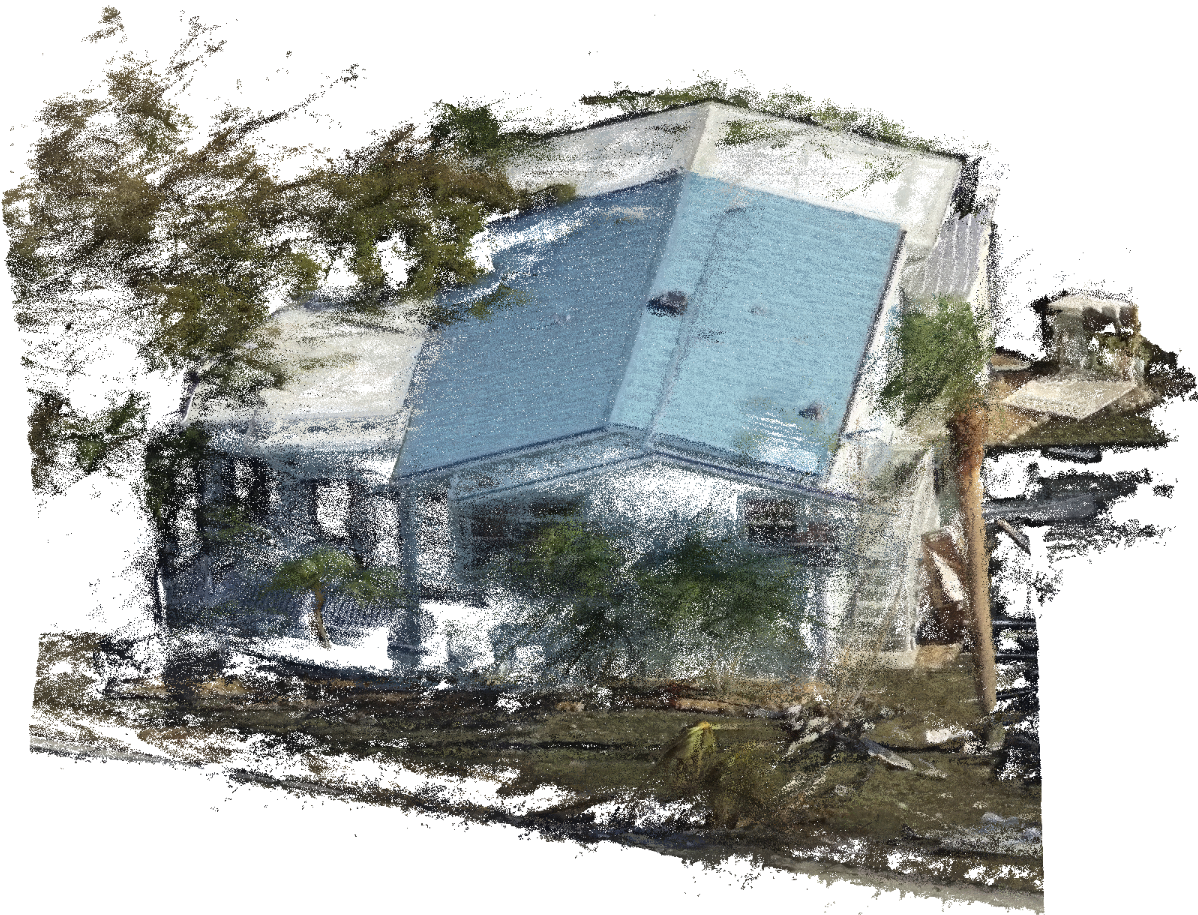} \\ 
Ground Truth & \includegraphics[width=0.2\linewidth]{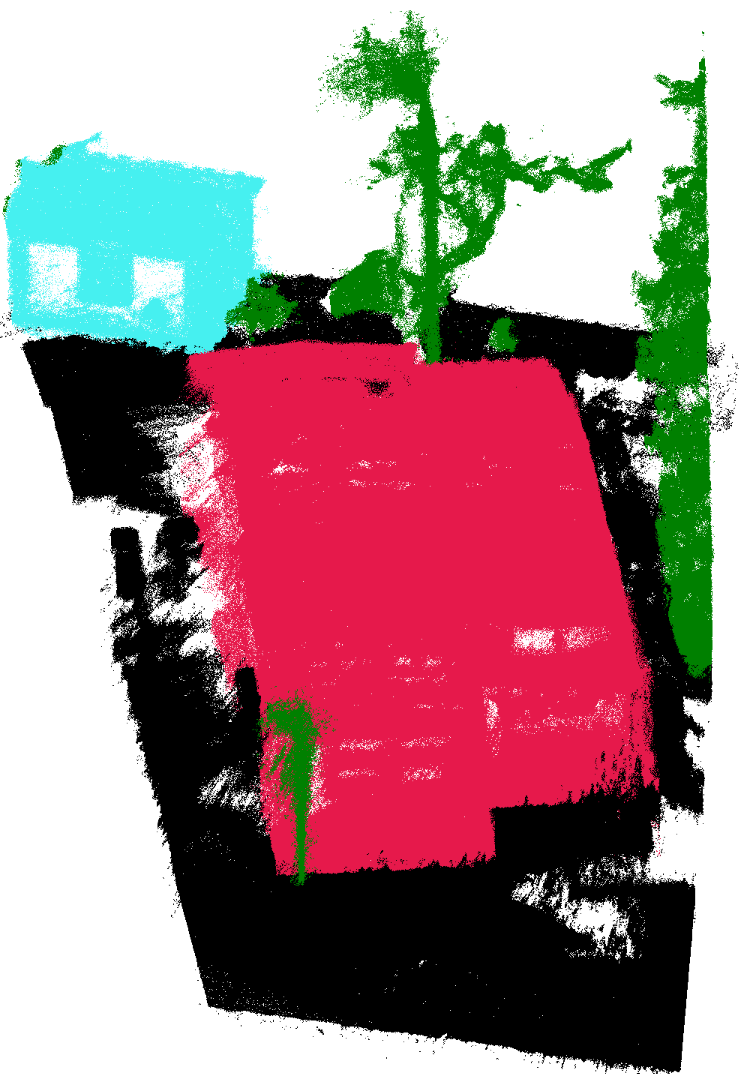}  & \includegraphics[width=0.3\linewidth]{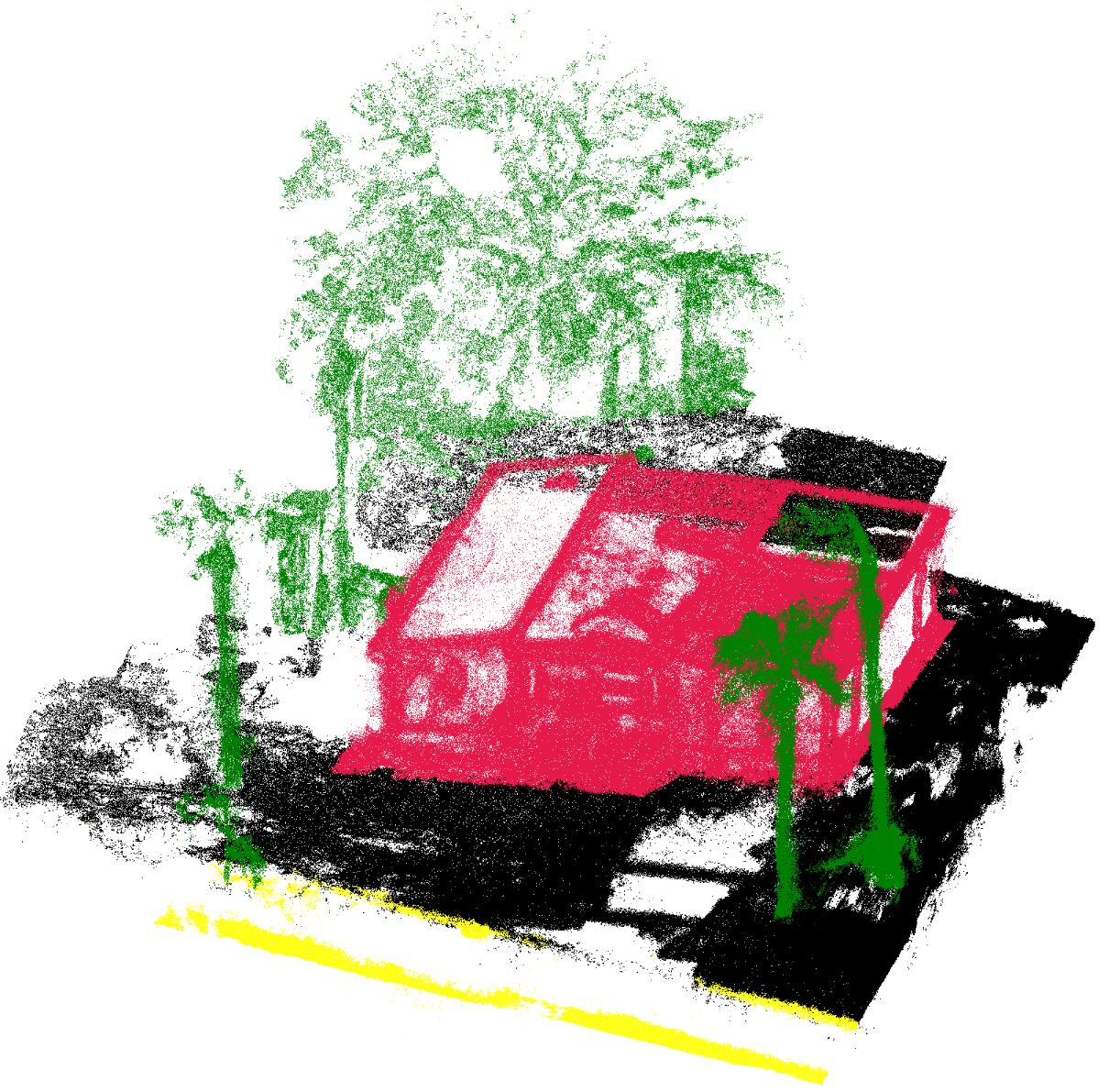} & \includegraphics[width=0.23\linewidth]{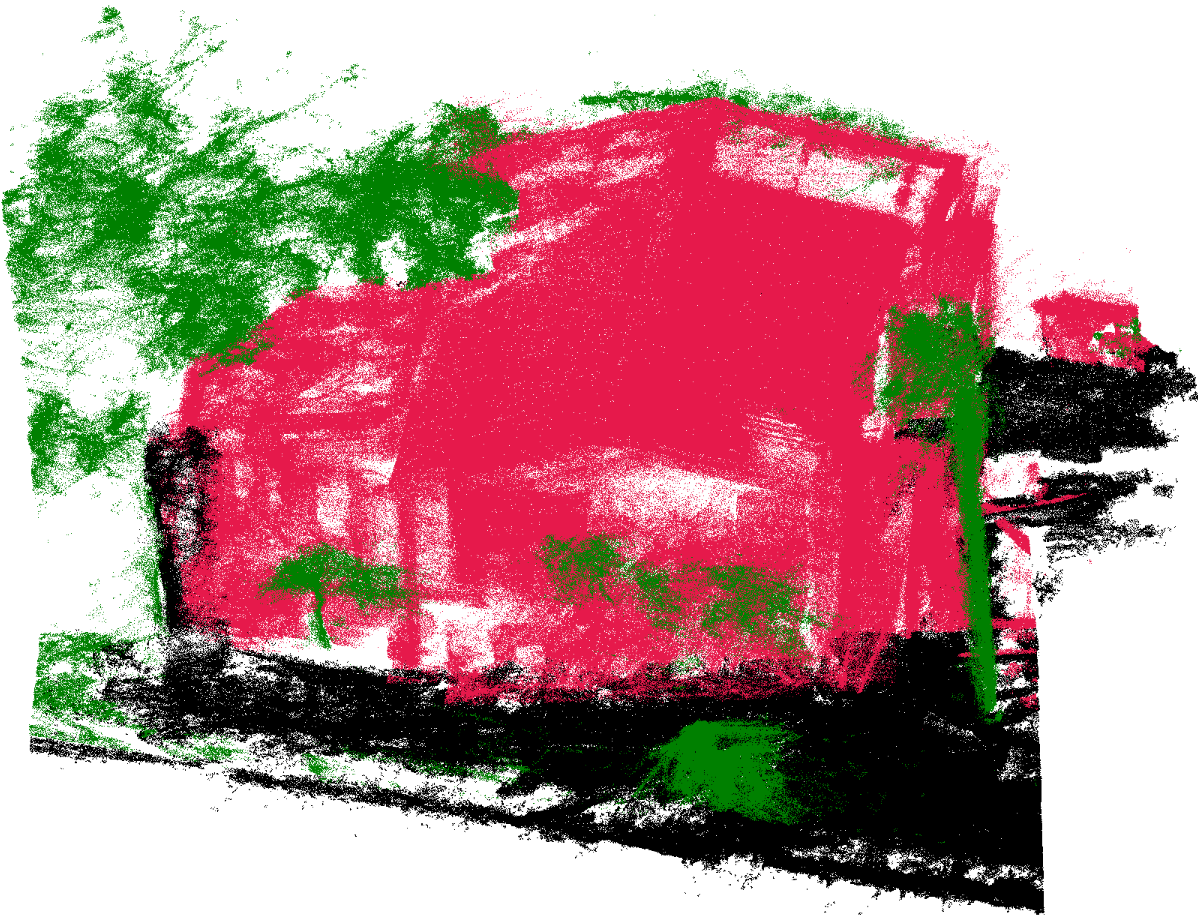}  \\
OA-CNNs \cite{Peng_2024_CVPR} & \includegraphics[width=0.2\linewidth]{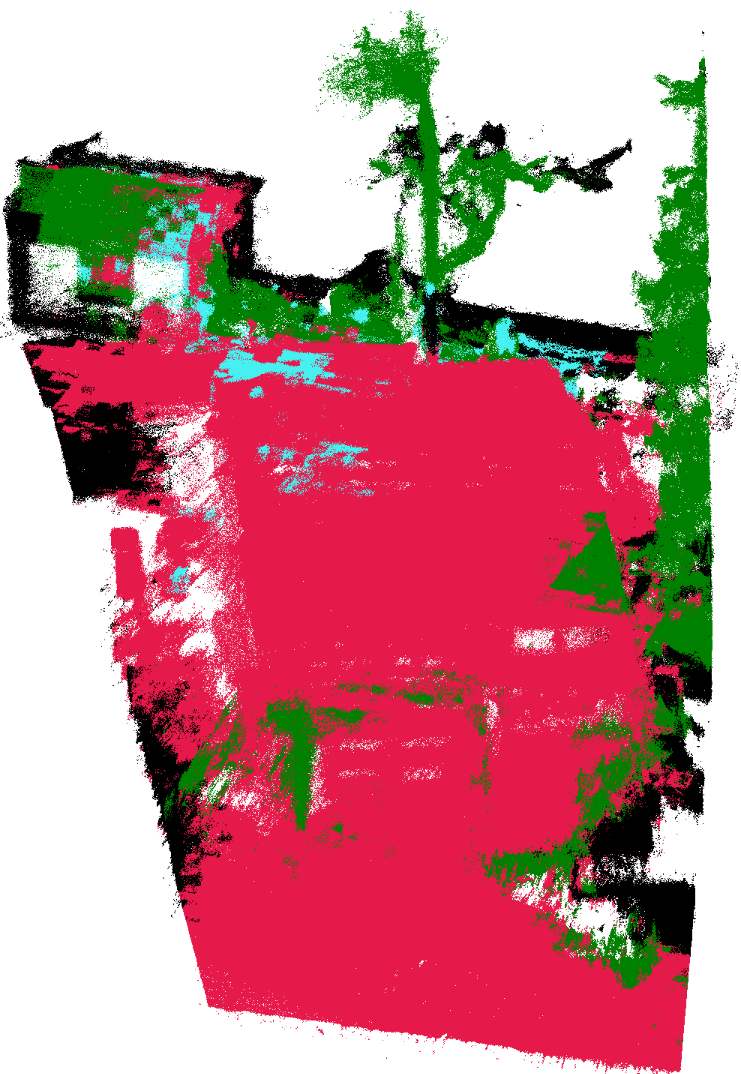}  & \includegraphics[width=0.3\linewidth]{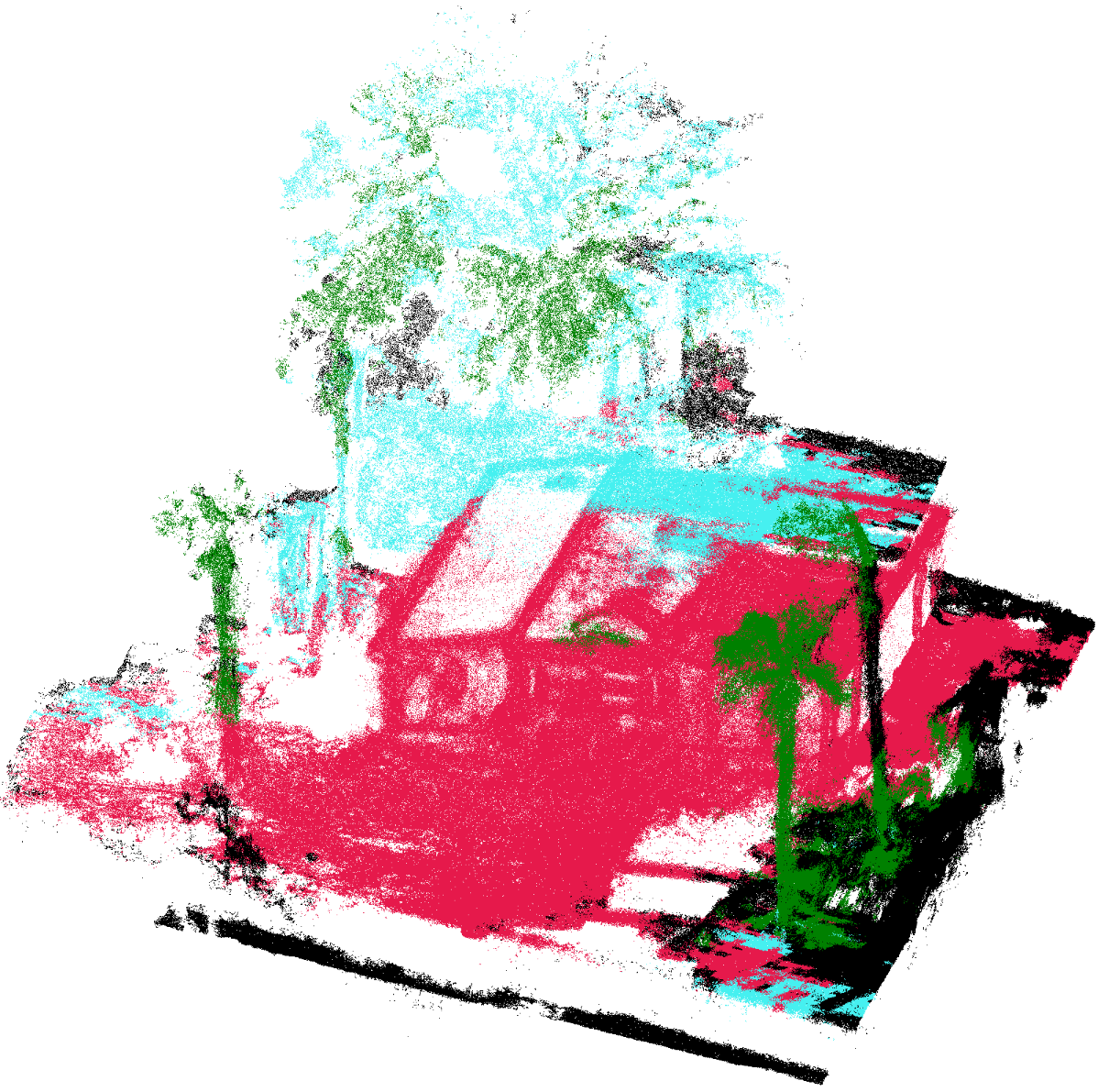} & \includegraphics[width=0.23\linewidth]{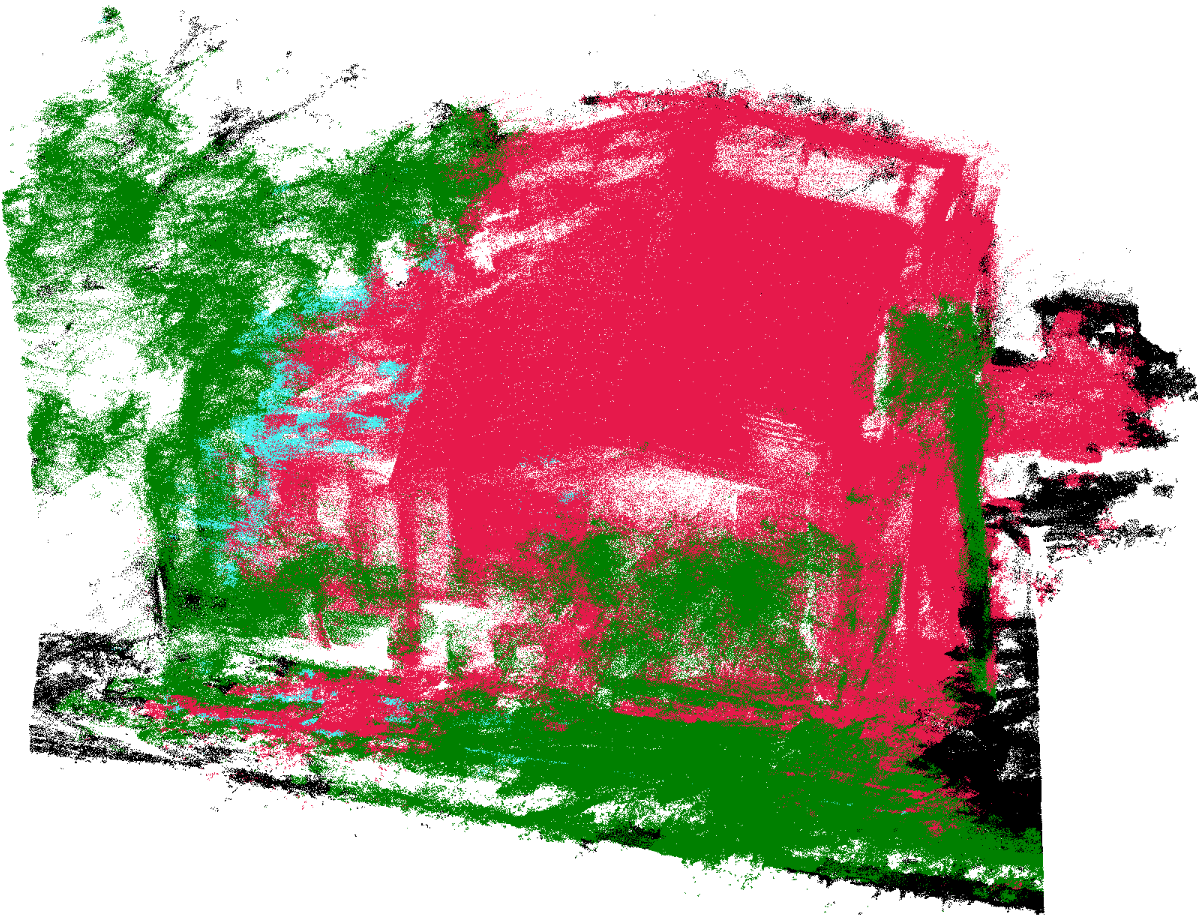}  \\ 
PTv3 \cite{wu2024ptv3} & \includegraphics[width=0.2\linewidth]{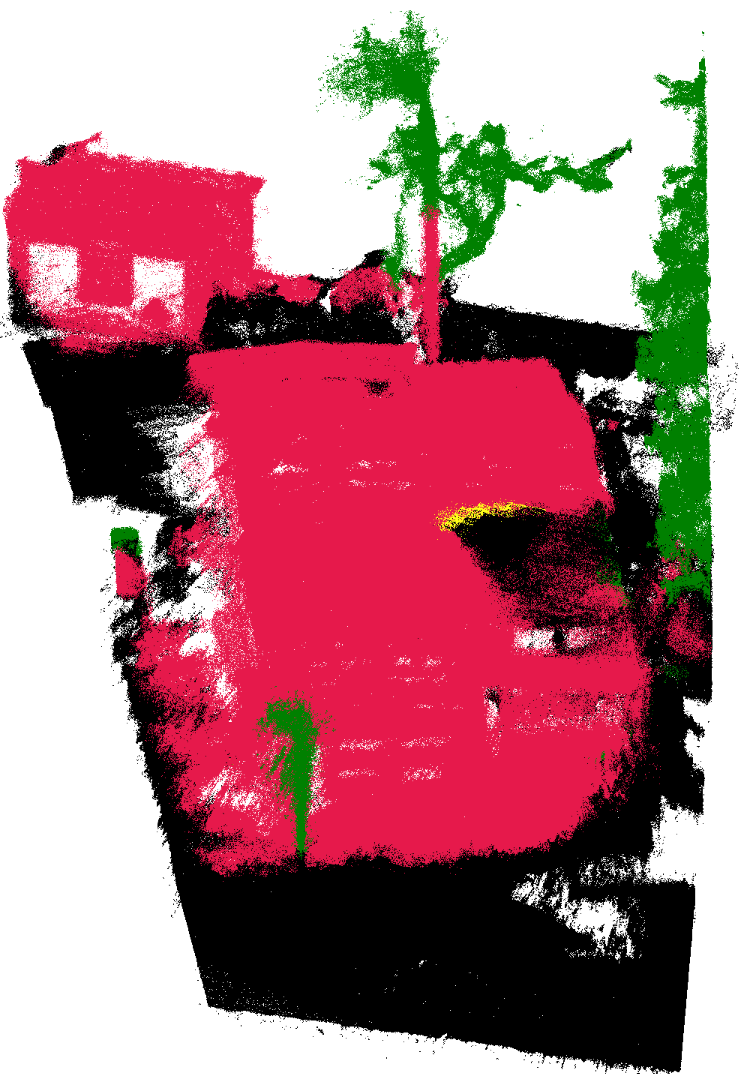}  & \includegraphics[width=0.3\linewidth]{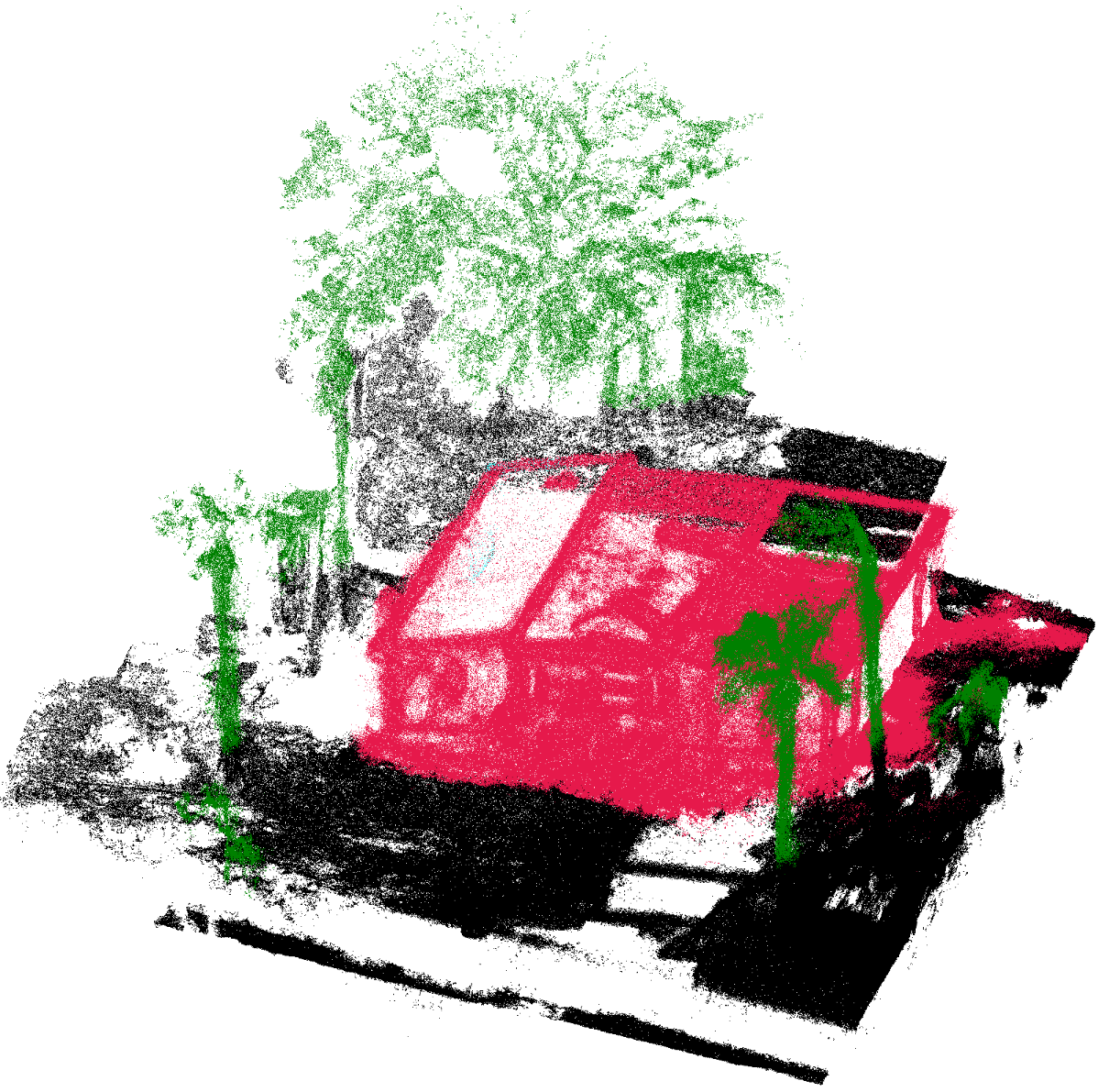} & \includegraphics[width=0.23\linewidth]{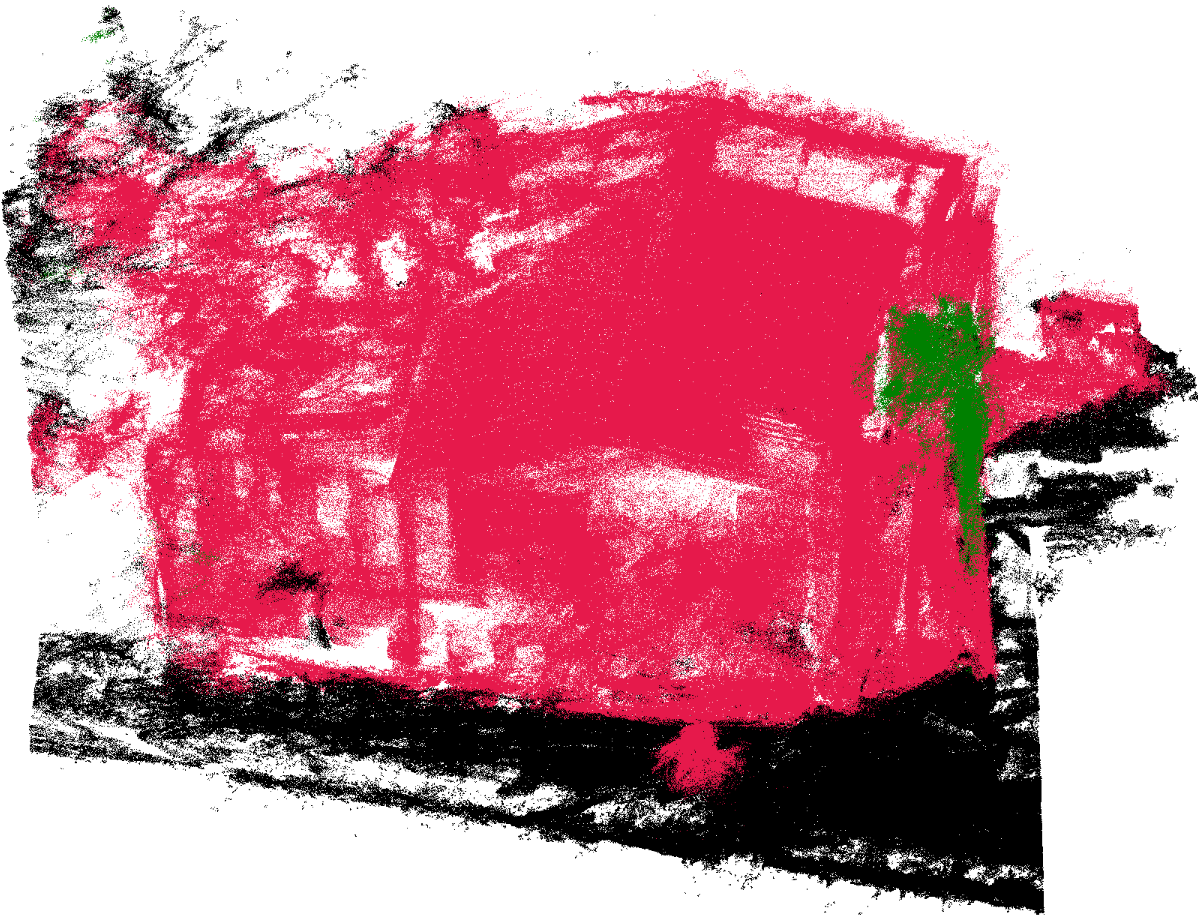}  \\ 
OPTNet (Ours) & \includegraphics[width=0.2\linewidth]{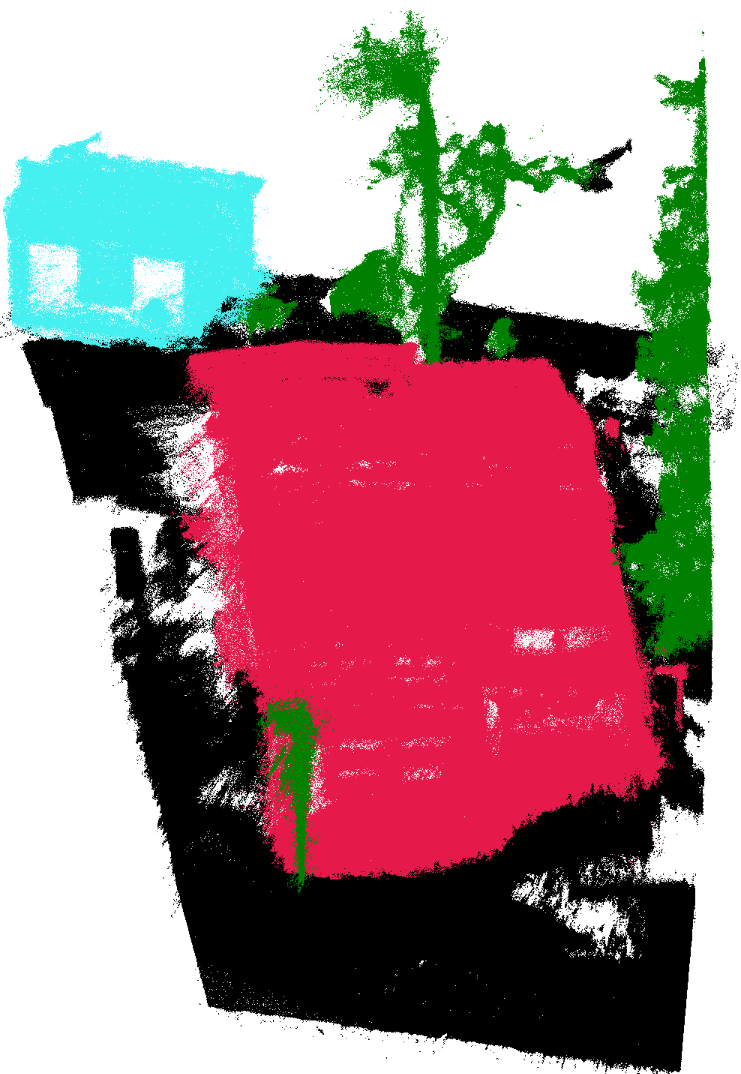}  & \includegraphics[width=0.3\linewidth]{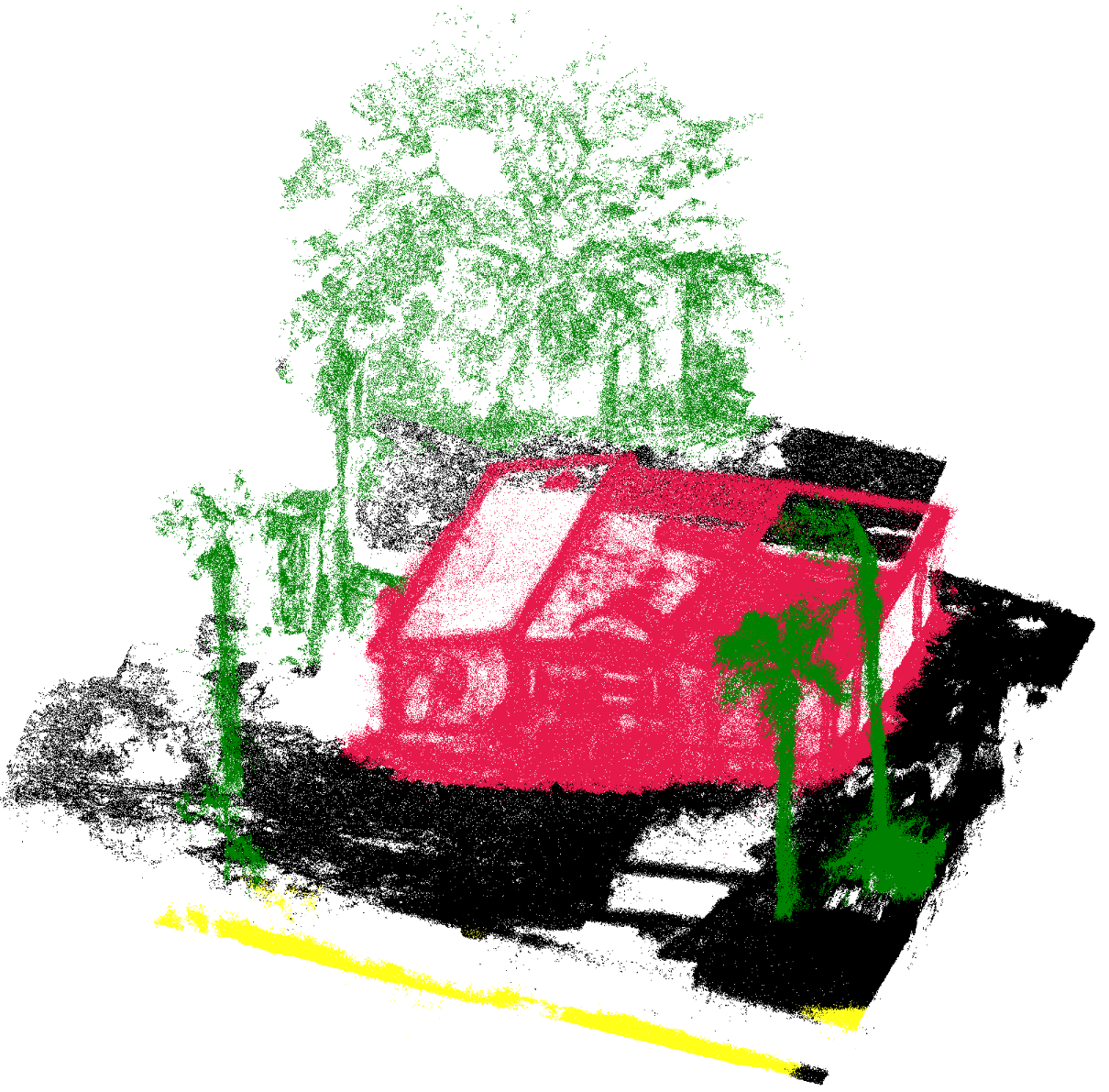} & \includegraphics[width=0.23\linewidth]{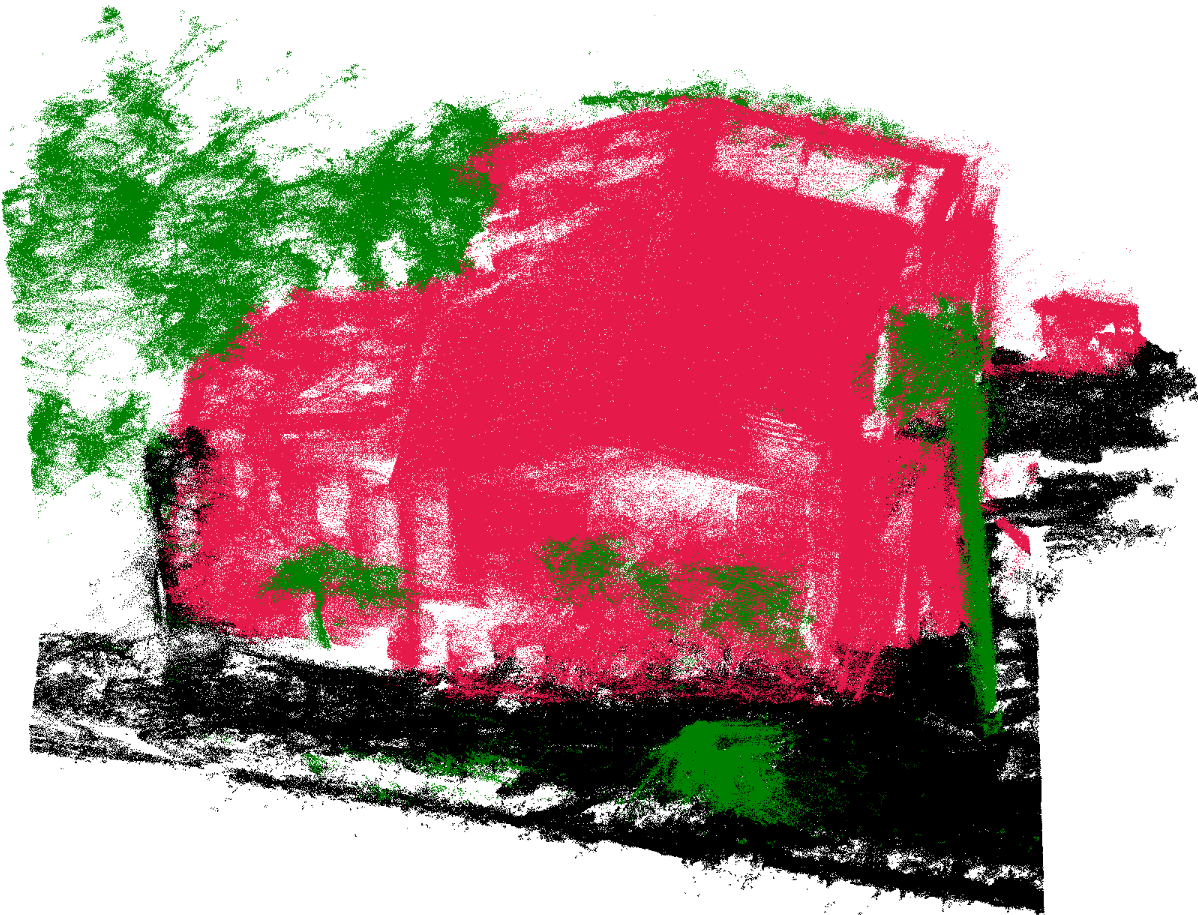}  \\ 
\end{tabular}
}
\label{tab:qual}
\end{table}
}
\subsection{Implementation Details}
OPTNet is implemented based on PTv3 \cite{wu2024ptv3} framework on 2 NVIDIA H100 GPUs.
We use the AdamW optimizer with a max learning rate of 0.006 and a OneCycleLR scheduler.
The models are trained for 3000 epochs on 3DAeroRelief. For the first 5 epochs (warm-up), we force the network to sort points using the shuffle from space filling curves (e.g., Z-order, Hilbert) to stabilize the transformer weights. Subsequently, we switch to the learned \emph{Point Sorter} for dynamic serialization.

\subsection{Quantitative Results}
\noindent\textbf{Results on 3DAeroRelief.} 
This dataset poses extreme challenges for existing methods due to the sheer scale and irregularity of UAV-derived point clouds.
Table~\ref{tab:iou_results} presents the Intersection over Union (IoU) comparison. Standard serialization-based methods like PTv3 \cite{wu2024ptv3} struggle to maintain feature coherence, achieving only 45.84\% mIoU.
Specifically, PTv3 \cite{wu2024ptv3} fails to capture the \textit{Road} class (4.67\% IoU), likely because the thin, elongated geometry of roads is broken by the static Hilbert curve ordering.
In contrast, OPTNet \cite{Peng_2024_CVPR} achieves a state-of-the-art 79.65\% mIoU, outperforming the strongest baseline (PTv3) by over 33 percentage points.
The improvement is consistent across all categories but is most dramatic in the \textit{Road} class (51.31\%), proving that our learnable sorter successfully adapts to elongated topologies that break static curves.
Table~\ref{tab:acc_results} details the Class Accuracy. OPTNet achieves the highest performance across all categories, including critical classes like \textit{Building-Damage} (97.47\%), where it surpasses OA-CNNs \cite{Peng_2024_CVPR} (92.76\%). While OA-CNNs shows competitive results on damaged buildings, it lacks robustness, completely failing to classify roads (0.00\%). In contrast, OPTNet delivers consistently superior performance, achieving >84\% accuracy across every class and a Mean Accuracy (mAcc) of 92.14\%, significantly outperforming the next best method (PTv3 \cite{wu2024ptv3} at 55.08\%).
\subsection{Qualitative Results}
We visualize the segmentation results in Table~\ref{tab:qual}. The ground truth and predictions for OA-CNNs \cite{Peng_2024_CVPR},  PTv3 \cite{wu2024ptv3}, and OPTNet are compared.
OPTNet demonstrates superior boundary preservation and classification accuracy, particularly for damaged buildings (shown in Red) and roads (Yellow), where other methods often produce fragmented or noisy outputs.
\section{Ablation Studies}
\label{sec:ablation}
\begin{table}[!t]
\centering
\caption{Ablation Study: \textbf{Intersection over Union (IoU)} for varying neighbor counts $k$. The selected setting ($k=24$) is highlighted.}
\label{tab:ablation_iou}
\begin{tabular}{c|c|c|c|c|c|c}
\toprule
\textbf{$k$} & \textbf{mIoU (\%)} & \textbf{Background} & \textbf{Bldg-Dmg} & \textbf{Bldg-No-Dmg} & \textbf{Road} & \textbf{Tree} \\
\midrule
4  & 78.10 & 92.02 & 93.55 & 72.11 & 48.74 & 84.06 \\
8  & \textbf{79.75} & 92.34 & 93.53 & 74.88 & \textbf{54.76} & 83.22 \\
16 & 79.47 & \textbf{92.55} & 93.99 & 74.53 & 51.55 & \textbf{84.76} \\
\textbf{24} & 79.65 & 92.40 & \textbf{94.60} & \textbf{77.08} & 51.31 & 82.85 \\
32 & 77.82 & 92.5 & 93.72 & 70.27 & 47.60 & 84.98 \\
\bottomrule
\end{tabular}
\end{table}
\begin{table}[!t]
\centering
\caption{Ablation Study: \textbf{Accuracy (Acc)} and \textbf{Overall Accuracy (OA)} for varying neighbor counts $k$. The selected setting ($k=24$) yields the highest Overall Accuracy.}
\label{tab:ablation_acc}
\begin{tabular}{c|c|c|c|c|c|c|c}
\toprule
\textbf{$k$} & \textbf{mAcc (\%)} & \textbf{OA (\%)} & \textbf{Background} & \textbf{Bldg-Dmg} & \textbf{Bldg-No-Dmg} & \textbf{Road} & \textbf{Tree} \\
\midrule
4  & 91.77 & 95.32 & 95.45 & 96.60 & 83.41 & 90.89 & 92.48 \\
8  & 92.23 & 95.46 & \textbf{96.10} & 96.00 & 87.23 & 89.50 & 92.32 \\
16 & \textbf{92.32} & 95.65 & 95.99 & 96.66 & 85.92 & \textbf{90.91} & 92.11 \\
\textbf{24} & 92.14 & \textbf{95.75} & 94.82 & \textbf{97.47} & \textbf{89.63} & 84.80 & \textbf{93.99} \\
32 & 90.33 & 95.47 & 96.56 & 96.44 & 81.82 & 87.28 & 89.55 \\
\bottomrule
\end{tabular}
\end{table}
We conduct ablation studies on the 3DAeroRelief dataset to validate the design choices of OPTNet. Specifically, we analyze the impact of the neighbor count hyperparameter $k$ used in our Self-Supervised Ordering Loss.
\\
The Locality Loss $\mathcal{L}_{local}$ relies on a local neighborhood of size $k$ to enforce geometric consistency in the learned serialization. We investigate the effect of varying $k \in \{4, 8, 16, 24, 32\}$.
Tables \ref{tab:ablation_iou} and \ref{tab:ablation_acc} present the Intersection over Union (IoU) and Accuracy metrics, respectively. \\
As observed, increasing $k$ initially improves performance by allowing the model to capture broader local context. The performance peaks at $k=24$, achieving high structural classification results (e.g., 94.60\% IoU for \textit{Building-Damage}). \\
However, increasing $k$ further to 32 leads to a performance drop (mIoU decreases to 77.82\% and mAcc to 90.33\%). This suggests that an overly large neighborhood forces the serialization to cluster distant, semantically unrelated points, thereby degrading the local receptive field quality (over-smoothing).
Although $k=8$ yields a marginally higher mIoU (79.75\%), $k=24$ offers the best Overall Accuracy (95.75\%) and significantly stronger performance on the critical building classes. Therefore, we select $k=24$ as the optimal setting to balance context aggregation and geometric fidelity.

\section{Conclusion}
\label{sec:conclusion}
In this paper, we introduced OPTNet, a novel framework that addresses the geometric limitations of static serialization in 3D semantic segmentation. Unlike traditional methods that rely on fixed space-filling curves, OPTNet employs a Learnable Point Sorter trained via a Self-Supervised Ordering Loss to dynamically optimize point traversal. This data-driven serialization ensures that spatially coherent points are grouped efficiently, maximizing the receptive field of windowed attention mechanisms. 
Our approach seamlessly replaces heuristic sorting in architectures like PTv3 \cite{wu2024ptv3}, effectively mitigating the boundary artifacts inherent in fixed orderings. 
Extensive evaluations on the 3DAeroRelief \cite{le20253daerorelief3dbenchmarkuav} dataset demonstrate that OPTNet significantly outperforms existing baselines, establishing a new state-of-the-art for post-disaster assessment and offering robust performance in the critical task of identifying damaged infrastructure.

%
%
%
\bibliographystyle{splncs04}
\bibliography{refs}
%
\end{document}